\documentclass{article}

\PassOptionsToPackage{square,numbers,compress}{natbib}
\usepackage[main,final]{neurips_2026}

\usepackage[utf8]{inputenc} 
\usepackage[T1]{fontenc}    
\usepackage{hyperref}       
\usepackage{url}            
\usepackage{booktabs}       
\usepackage{amsfonts}       
\usepackage{nicefrac}       
\usepackage{xcolor}         

\usepackage{graphicx}
\usepackage{amsmath}
\usepackage{enumitem}
\usepackage{needspace}
\usepackage{xspace}         

\title{NestedVLA: Learning to Consolidate and Generate Skills for Vision-Language-Action Models}

\author{%
  Guodong Du \qquad Wanyu Lin\thanks{Corresponding author.}\\
  Department of Computing, The Hong Kong Polytechnic University\\
  \texttt{guodong.du@connect.polyu.hk} \quad \texttt{wan-yu.lin@polyu.edu.hk}
}

\begin{document}
\maketitle
\begin{abstract}
Vision--language--action (VLA) models can learn a wide range of robotic
skills, but combining separately trained skills remains challenging.
The original training data may be unavailable for joint retraining,
while directly merging model weights can cause interference.
We propose \textbf{NestedVLA}, which consolidates independently
fine-tuned VLA policies without access to their original training data.
Instead of merging all updates into one checkpoint, a hypernetwork
embedded in the VLA generates action-expert weight updates for a
selected task. Each task's compressed vision--language model update
is stored separately, and both updates are added to a frozen base VLA
to instantiate the corresponding policy.
We evaluate NestedVLA by consolidating three policies fine-tuned with
reinforcement learning on MetaWorld, ManiSkill, and CALVIN.
It achieves $66.7\%$ average success, improving on the strongest
model-merging baseline by $41.1$ percentage points.
It performs within $1$ percentage point of per-task fine-tuning on
ManiSkill and substantially narrows the gap on CALVIN.
In this three-policy setting, total storage is $1.5\times$ that of
a single full checkpoint. Code is available at
\url{https://github.com/WanyuGroup/NeurIPS2026-NestedVLA}.
\end{abstract}

\section{Introduction}
\label{sec:intro}

Vision-language-action (VLA) models combine visual and language understanding
with embodied control~\citep{rt1,rt2,OpenVLA,pi0,pi0.5}. Independently
fine-tuning a common backbone produces policies specialized for different
robotic benchmarks, but storing one full checkpoint per benchmark duplicates
most parameters. Joint multi-task training can avoid this duplication when
data from all benchmarks can be pooled. We instead study post-hoc
consolidation of policies independently fine-tuned from the same backbone.
We seek to consolidate their task-specific parameter updates into a compact
representation without pooling the original training data.

\begin{figure}[t]
\centering
\includegraphics[width=\linewidth]{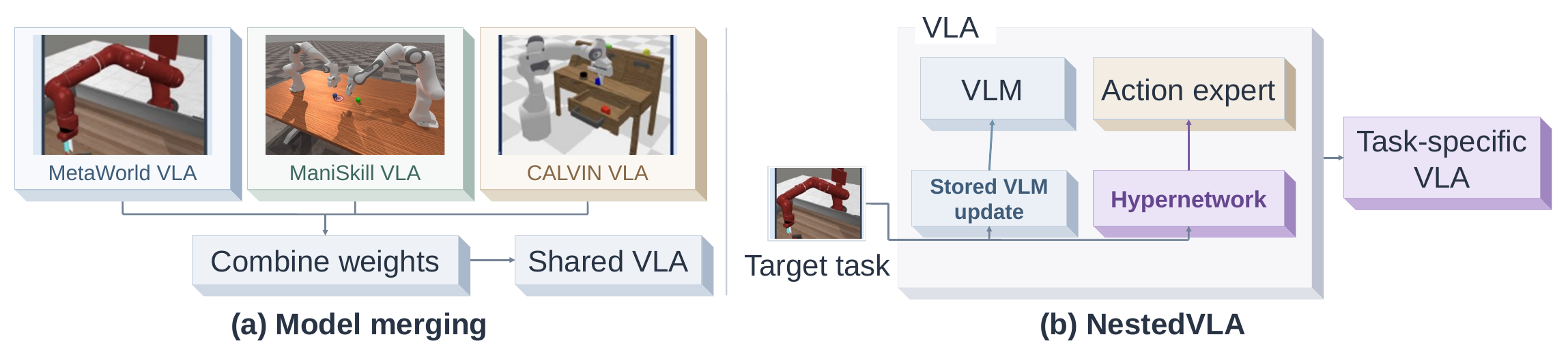}
\caption{\textbf{Motivation and overview of NestedVLA.}
(a) Model merging combines independently fine-tuned VLA weights into a
shared model, where incompatible updates can impair the source tasks.
(b) NestedVLA retains a compressed task-specific VLM update and uses an
embedded task-conditioned hypernetwork to generate the action-expert update.
Both updates are applied to a shared base to materialize the selected VLA.}
\label{fig:background}
\end{figure}

A natural post-hoc approach is to merge the parameter updates of the
fine-tuned policies into a single checkpoint. In VLA models, however,
independently learned updates can interfere strongly when combined: in our
experiments, simple averaging loses most of the performance of the separate
policies. Our weight-space analysis finds weak alignment across major
parameter groups, an incompatibility that is particularly consequential in
the action expert because it directly governs control
(Appendix~\ref{app:conflict_diag}). This observation is consistent with
MergeVLA~\citep{fu2025mergevla}, which likewise identifies dependencies
distributed across action-expert layers as a major obstacle to cross-skill
merging. These findings motivate a consolidation strategy that preserves
policy-specific parameter structure rather than collapsing all updates into
one set of weights.

We propose \textbf{NestedVLA}, a framework that shifts the paradigm from
parameter aggregation to parameter generation. A hypernetwork embedded in
the VLA generates sparse action-expert updates conditioned on the target
task, while compressed task-specific VLM residuals are retained separately.
The selected updates compose with a shared base to instantiate a
task-specific VLA without changing its forward architecture or introducing
routing supervision. We evaluate NestedVLA on MetaWorld~\citep{mcleanmeta},
ManiSkill~\citep{maniskill}, and CALVIN~\citep{calvin} using the open-source
$\pi_{0.5}$~\citep{pi0.5} backbone. NestedVLA outperforms the strongest
model-merging baseline by $41$\,pp on average, matches per-task fine-tuning
within $1$\,pp on ManiSkill, and substantially closes the gap on CALVIN.

\section{Related Work}
\label{sec:related_work}

\paragraph{Vision-Language-Action models.}
Recent VLAs cast robotic control as instruction-following on top of
pretrained vision-language backbones~\citep{rt1, rt2, OpenVLA, prismatic,
pi0, pi0.5}. OpenVLA~\citep{OpenVLA} discretizes actions into rare
vocabulary tokens for autoregressive prediction; $\pi_0$~\citep{pi0}
pairs the VLM with a separate \emph{action expert} trained via flow
matching for continuous action chunks; $\pi_{0.5}$~\citep{pi0.5}
extends this design to interleave sub-task narration and low-level
control. Lighter variants such as VLA-Adapter~\citep{vla-adapter}
preserve multi-task competence at sub-billion-parameter scales. VLAct
uses representation-centric continued pre-training to preserve VLM priors
and promote shared action semantics across
embodiments~\citep{yang2026vlact}. Across this family, multi-task ability
is learned from pooled data, requiring co-located datasets and predefined
skills. We target the regime where these assumptions break: skills are
acquired \emph{independently} and consolidated \emph{after the fact}.

\paragraph{Parameter-efficient visual adaptation.}
Prompt- and adapter-based methods offer lightweight alternatives to full
fine-tuning of vision backbones. Recent work makes prompts instance-aware,
injects structured visual semantics, or searches for layer-specific prompt
fusion rules~\citep{xiao2025visual,xiao2026adapting,xiao2026layer}.
Structured low-rank adaptation further filters task-irrelevant directions
and coordinates updates across layers~\citep{xiao2026not}, while visual
alignment methods address over-reliance on language priors in multimodal
models~\citep{xiao2026staying}. \citet{meng2026ahra} propose AHRA, a
hierarchical shared--private expert framework that aligns semantic levels
while preserving modality-specific evidence. Such methods adapt or organize
multimodal representations; NestedVLA instead starts from independently trained
VLA policies and consolidates their parameter updates post hoc.

\paragraph{Hypernetworks.}
Using one network to predict the parameters of another dates back to
HyperNetworks~\citep{ha2017hypernetworks}, with later applications to
continual learning~\citep{vonoswald2020continual} and multi-task
NLP~\citep{mahabadi2021parameter}. Recent parameter generators use latent
diffusion to synthesize neural-network weights~\citep{wang2024pdiff}, scale
this process to full models with recurrent generation~\citep{wang2025rpg},
map task prompts directly to LoRA weights~\citep{liang2025dnd}, or produce
instance-conditioned weight updates as functional memory~\citep{wu2026hywu}.
NestedVLA inherits this task-conditioned generation premise, but targets a
sparse top-$K$ delta on a frozen billion-scale VLA backbone via patch-based
tokenization, and consolidates independently fine-tuned RL policies rather
than learning skills from scratch.

\paragraph{Reusable task updates.}
Related work on language models compresses full-model task updates into
modular SkillPacks for heterogeneous knowledge fusion and multi-domain
specialization~\citep{du2026graftllm,li2026skillweave}. HoE instead extracts
LoRA-form experts from task vectors and routes them for multi-objective
alignment~\citep{li2026hoe}. These studies concern text-only language models
and retain or select task modules. NestedVLA studies closed-loop VLA policies
and generates the action-expert update with an embedded hypernetwork.

\paragraph{Model merging.}
Model merging combines independently trained checkpoints into a single
deployable model. Recent surveys organize model fusion across parameter,
representation, and behavior levels~\citep{cai2026fusion,
zhou2025democratizing,zhou2026model}. \textbf{Static merging} produces one fixed delta via
weighted averaging~\citep{modelsoup, fishermerging, TA, du2024knowledge},
conflict-handling~\citep{ties, dare, wudi, pcb, fang2025see}, or low-rank
sub-spaces~\citep{tsvm, knots, iso-cts, twin, nps}, with adapter
counterparts~\citep{pem-composition, do-merging}. \textbf{Dynamic
merging} retains per-task masks, scalars, residuals, or
frequency-domain corrections at inference~\citep{tall-mask, emr,
free-merging, mmer, didi2026}. Recent adaptive methods build reusable
component-level expert libraries for input-aware recombination~\citep{qiu2026merge}
or evolve experts with manifold-aware, training-free routing for continual
merging~\citep{qiu2026madeit}. Across both regimes the operative verb is
\emph{select}: every output value is an interpolation, mask, or
projection of values already present in $\{\theta_m-\theta_0\}_m$.
Complementary studies make expert training merge-aware~\citep{gu2026smat},
calibrate or quantize post-merge models~\citep{gu2026featcal,wang2026epmq},
and analyze scaling laws or expert-read budgets for weight-space
merging~\citep{wang2026modelmergingscaling,wang2026mergepipe}.
NestedVLA \emph{generates} the per-skill delta from a single shared
hypernetwork instead. Embodied applications of merging remain largely
unexplored; ReVLA~\citep{revla} mitigates single-skill domain
forgetting rather than multi-skill consolidation.

\section{NestedVLA}
\label{sec:method}

\subsection{Problem Formulation}
\label{sec:method_formulation}

We consider $M$ task-specific datasets
$\mathfrak{D}=\{\mathcal{D}_m\}_{m=1}^{M}$, each corresponding to a
manipulation task in a benchmark-specific setting. Fine-tuning
a pretrained VLA backbone $\Theta_0$ on each $\mathcal{D}_m$ yields
task-specific parameters $\{\Theta_m\}_{m=1}^{M}$ and the corresponding
\emph{task vector}~\citep{TA}
\begin{equation}
\tau_m \;=\; \Theta_m - \Theta_0,
\end{equation}
which captures all parameter changes induced by task $m$. We treat each
$\tau_m$ as the parametric encoding of a \emph{skill}, a reusable
manipulation capability acquired in isolation. The goal of skill
consolidation is to construct a unified model that leverages the entire
skill set $\{\tau_m\}_{m=1}^{M}$ across diverse downstream scenarios
without storing full per-task checkpoints or using expert routing.

A common approach is to merge $\{\tau_m\}$ into a single static
delta~\citep{TA, ties, modelsoup, fishermerging, wudi}.
However, parameter aggregation in the VLA setting suffers from severe
inter-task conflicts (Sec.~\ref{sec:intro}), as different tasks impose
contradictory updates on the same parameters. We instead propose to
\emph{generate} parameters. Write
$\Theta_m=(\Theta_m^V,\Theta_m^A)$ for the VLM and action-expert
parameters. In the evaluated v1 system, a generator $\mathcal{G}_\phi$,
conditioned on task identifier $m$, produces the action-expert delta,
while a compressed VLM delta $z_m^V$ is retained for the same task:
\begin{equation}
\hat{\tau}_m^A=\mathcal{G}_\phi(m),\qquad
\hat{\Theta}_m^A=\Theta_0^A+\hat{\tau}_m^A,\qquad
\hat{\Theta}_m^V=\Theta_0^V+z_m^V.
\label{eq:generator}
\end{equation}
Both updates compose with the frozen base at inference. This shifts the
action-side problem from \emph{parameter aggregation} to
\emph{parameter generation}, avoiding full per-task checkpoints and a
single conflicting merge.

\subsection{Overview}
\label{sec:method_overview}

\begin{figure}[t]
\centering
\includegraphics[width=\linewidth]{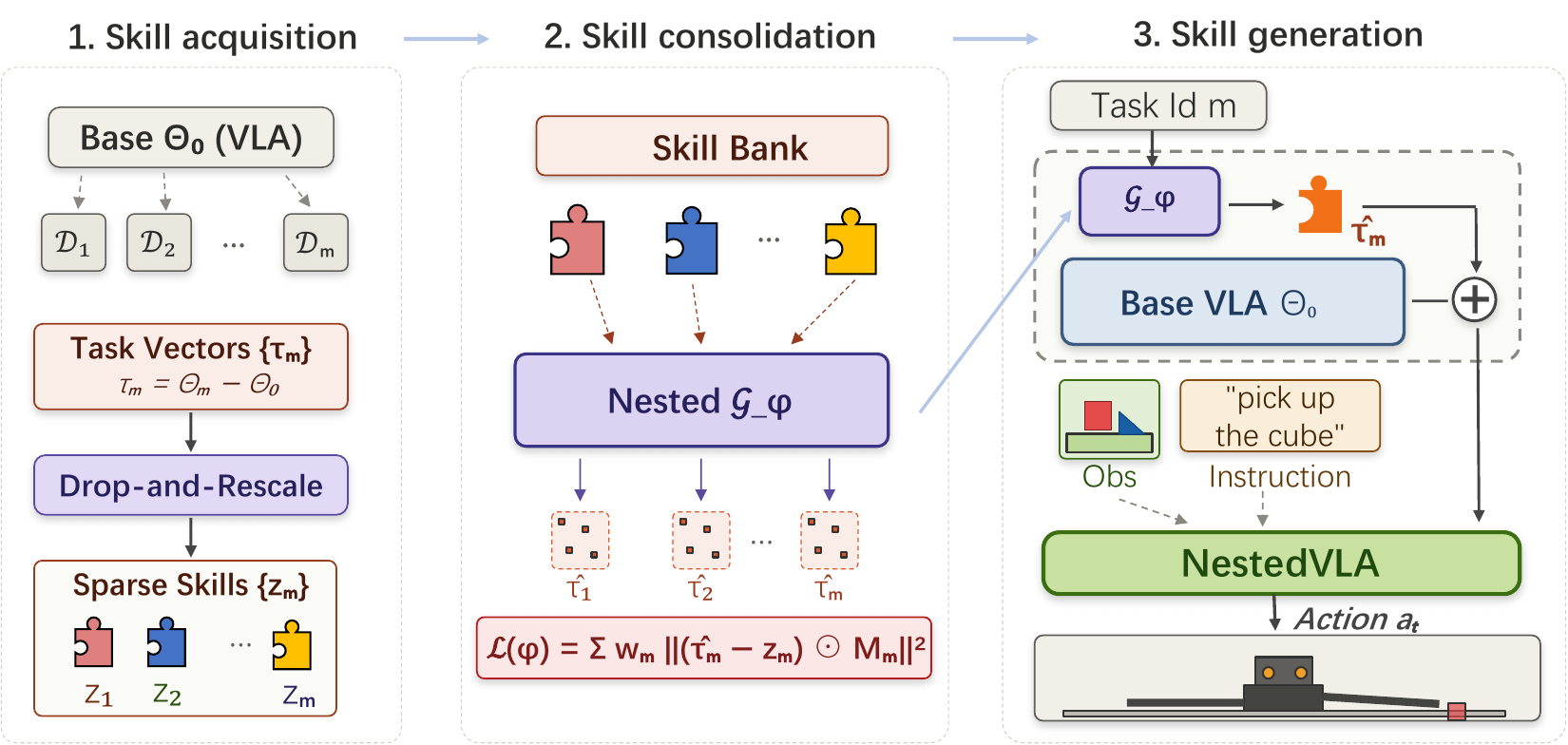}
\caption{\textbf{NestedVLA pipeline.} (1) \emph{Skill acquisition}:
each task is fine-tuned independently from the same base $\Theta_0$
and the resulting task vector $\tau_m$ is compressed into a sparse
skill $z_m=\alpha\cdot\tau_m\odot M_m$ via drop-and-rescale.
(2) \emph{Skill consolidation}: a single nested hypernetwork
$\mathcal{G}_\phi$ is trained against the action-expert portions of
$\{z_m\}$ with a multi-task masked MSE loss.
(3) \emph{Skill generation}: at inference, querying $\mathcal{G}_\phi$
with task id $m$ generates the action-expert update; the corresponding
compressed VLM update is retrieved and both compose with the frozen base.}
\label{fig:method}
\end{figure}

NestedVLA realizes the generator $\mathcal{G}_\phi$ as a hypernetwork
\emph{nested} inside the VLA pipeline: it shares the backbone's
parameter layout as its output space, is trained against targets
derived from the backbone's task vectors, and is queried jointly with
the backbone at inference. The framework operates in three stages
(Fig.~\ref{fig:method}):

\begin{enumerate}
\item \textbf{Skill Acquisition} (\S\ref{sec:method_acquisition}):
  fine-tune $\Theta_0$ on each $\mathcal{D}_m$, then compress the
  resulting task vector $\tau_m$ into a sparse skill $z_m$ via
  drop-and-rescale.
\item \textbf{Skill Consolidation} (\S\ref{sec:method_consolidation}):
  train one nested hypernetwork $\mathcal{G}_\phi$ to absorb all $M$
  action-expert targets $\{z_m^A\}_{m=1}^{M}$ into shared parameters
  $\phi$ with task embeddings.
\item \textbf{Skill Generation} (\S\ref{sec:method_generation}):
  at inference, query $\mathcal{G}_\phi$ with the target task identity
  to generate the action-expert delta, retrieve the compressed VLM
  delta, and compose both updates with the frozen base.
\end{enumerate}

\subsection{Skill Acquisition}
\label{sec:method_acquisition}

The acquisition stage produces, for each task, a sparse skill
representation $z_m$ that the hypernetwork can later be trained against.
It comprises two steps: independent fine-tuning to obtain $\tau_m$, and
drop-and-rescale compression to turn $\tau_m$ into the compact $z_m$.

\paragraph{Independent fine-tuning.}
For each $\mathcal{D}_m$, we fine-tune $\Theta_0$ to obtain $\Theta_m$
and the task vector $\tau_m=\Theta_m-\Theta_0$. We make no assumption
on the training signal: $\Theta_m$ may be obtained via supervised
fine-tuning, reinforcement learning, or any combination thereof, as
long as all $\{\Theta_m\}$ are derived from the same $\Theta_0$. This
decoupling matches the data-silo realities of robotics, where
specialized skills are often trained independently in their native
pipelines.

\paragraph{Drop-and-rescale compression.}
A naive option is to use the dense $\tau_m$ as the hypernetwork target,
but this is prohibitive in both memory and learnability for VLA
backbones with billions of parameters. We observe that VLA fine-tuning
deltas exhibit a \emph{sparse-magnitude-dominated} structure: a small
fraction of elements concentrate the bulk of the delta's $\ell_2$
energy, while the rest behave like low-amplitude noise; in contrast,
LLM SFT deltas are typically denser and benefit from
quantization-plus-low-rank schemes such as D-QRELO~\citep{li2026d}. We
compared low-bit quantization, low-rank decomposition, and
drop-and-rescale on VLA deltas and found drop-and-rescale dominates
the fidelity-versus-compression trade-off
(\S\ref{app:compression}); we adopt it throughout NestedVLA.

\paragraph{Operator definition.}
Let $\tau_m^{(p)} \in \mathbb{R}^{d_p}$ denote the slice of $\tau_m$
for parameter tensor $p$. For a module budget $K \in (0,1]$, we compute
a per-module top-$K$ binary mask $M_m^{(p)}$ that retains the elements
with the largest magnitude within the module $p$ belongs to. The skill
representation is
\begin{equation}
z_m^{(p)} \;=\; \alpha \cdot \tau_m^{(p)} \odot M_m^{(p)},
\label{eq:drop_rescale}
\end{equation}
where $\alpha$ is a global rescale factor that compensates for the
energy lost during sparsification. Module-level (rather than
per-tensor) top-$K$ allocates the budget adaptively across layers,
which improves downstream fidelity (\S\ref{app:compression}). The
mask is deterministic given $\tau_m$: it is precomputed once and
applied multiplicatively to the loss at training (Eq.~\ref{eq:loss})
and to the head output at inference (Eq.~\ref{eq:inference}). The
hypernetwork only learns the dense values on $M_m{=}1$ positions.

\subsection{Skill Consolidation}
\label{sec:method_consolidation}

Given the action-expert targets $\{z_m^A\}_{m=1}^{M}$, the consolidation
stage trains a single hypernetwork $\mathcal{G}_\phi$ that absorbs the
entire skill set into one set of shared parameters $\phi$. We describe
the parameter tokenization that exposes the backbone layout to the
hypernetwork (\S\ref{sec:method_tokenization}), the architecture of
$\mathcal{G}_\phi$ (\S\ref{sec:method_arch}), and the training
objective (\S\ref{sec:method_loss}).

\subsubsection{Patch-based Parameter Tokenization}
\label{sec:method_tokenization}

A direct regression $\mathcal{G}_\phi(m) \to z_m$ over a billion-element
output is infeasible. We tokenize the parameter space into a 2D grid
of \emph{patches}. Each weight tensor in the backbone is split into
fixed-size $H_\text{patch}\times W_\text{patch}$ tiles (large tensors
span multiple patches; small tensors are zero-padded into one). Every
patch is uniquely addressed by four indices: parameter type (\texttt{ptype},
e.g.\ self-attention $Q$/$K$/$V$/$O$, FFN gate/up/down, layer norm),
\texttt{layer} (transformer block index), and \texttt{patch\_h},
\texttt{patch\_w} (tile coordinates within the source tensor). This
factorization exposes a structured spatial layout to the generator
while keeping the per-token output dimension bounded. Concrete patch
counts and tile sizes are reported in \S\ref{sec:exp_setup}.

\subsubsection{Hypernetwork Architecture}
\label{sec:method_arch}

\begin{figure}[t]
\centering
\includegraphics[width=\linewidth]{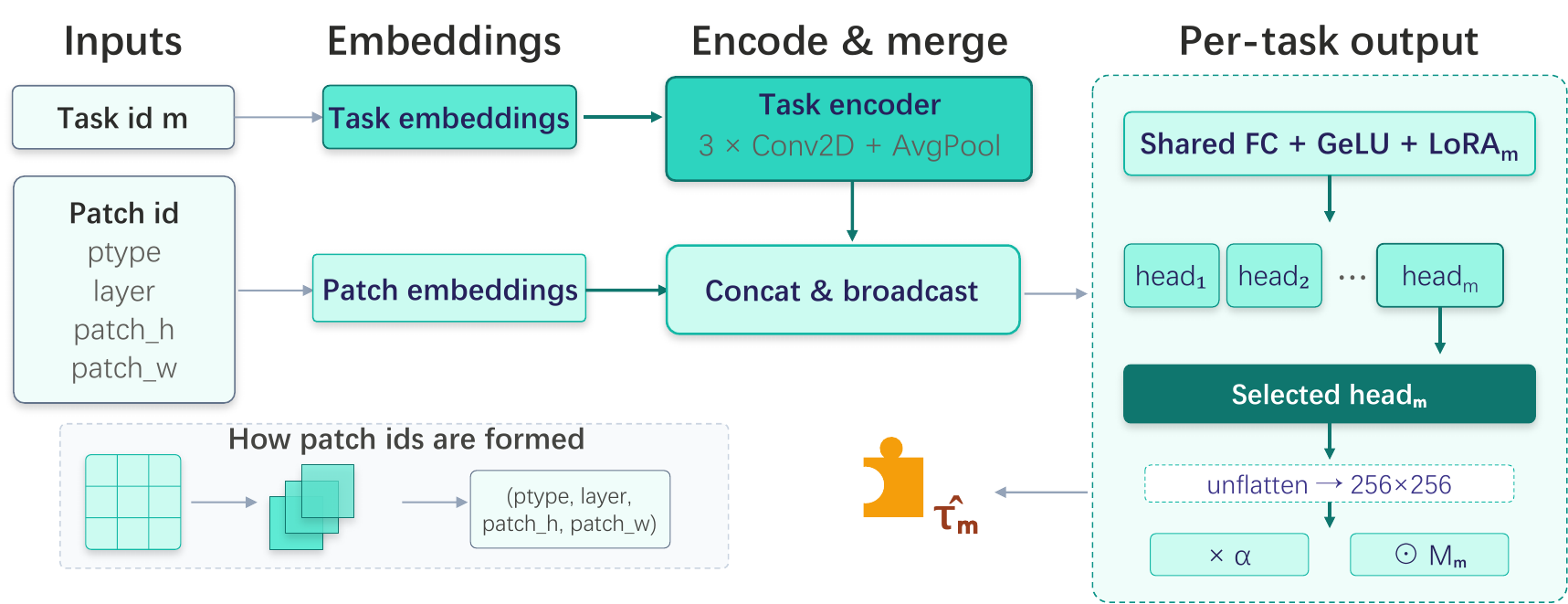}
\caption{\textbf{Architecture of the nested hypernetwork
$\mathcal{G}_\phi$.} Patches are addressed by
$(\texttt{ptype},\texttt{layer},\texttt{patch\_h},\texttt{patch\_w})$
and mapped to four per-task patch embeddings (\emph{patch path}). The
task id $m$ is mapped to a task embedding and processed by a
lightweight \emph{task encoder} (2D convolutions plus average pooling)
into a single task feature $f_\text{task}$. $f_\text{task}$ is
broadcast across all patches and concatenated with the patch
embeddings, then passed through a \emph{shared FC trunk} augmented
with a per-task low-rank LoRA path (\emph{LoRA path}). The trunk
output is decoded by a \emph{multi-head output} module that allocates
one fully-connected projection per task (\emph{multi-head path})
into a flat patch prediction. The precomputed action-expert mask
$M_m^A$ and global rescale $\alpha$ produce
$\hat{\tau}_m^A=\alpha\cdot\mathcal{G}_\phi(m)\odot M_m^A$; the
compressed VLM update is retained separately.}
\label{fig:arch}
\end{figure}

$\mathcal{G}_\phi$ has five trainable components, illustrated in
Fig.~\ref{fig:arch}: per-task \emph{patch embeddings}, a \emph{task
encoder}, a \emph{concat-and-broadcast} step, a \emph{shared FC trunk}
with a per-task LoRA path, and a \emph{multi-head output} module. In our
ablations (\S\ref{sec:exp_ablation}), the three task-conditioned
components (patch embeddings, LoRA, multi-head output) are called
the \emph{patch path}, \emph{LoRA path}, and \emph{multi-head path}.

\paragraph{Per-task patch embeddings.}
Each of the four patch indices (\texttt{ptype}, \texttt{layer},
\texttt{patch\_h}, \texttt{patch\_w}) is mapped through a learnable
embedding table indexed by both the patch index and the task id. The
four resulting vectors are concatenated to form a per-patch feature.
Per-task indexing (as opposed to a single shared table) lets the
hypernetwork use different effective patch addresses for different
tasks, which gives a small but consistent gain in our ablations.

\paragraph{Task encoder.}
The task identifier $m$ is mapped to a learnable task embedding
$\text{task\_emb}(m)$, reshaped onto a small spatial grid, and
processed by a lightweight 2D convolutional encoder followed by global
average pooling, yielding a single task feature $f_\text{task} \in
\mathbb{R}^{D_t}$. This vector is the only carrier of task identity
entering the rest of the network. The 2D-conv design is empirical; we
did not observe meaningful gains from increasing its depth or width.

\paragraph{Concatenate and broadcast.}
$f_\text{task}$ is broadcast across all patches and concatenated with
the four patch embeddings to form a per-patch input
$[f_\text{task};\, \texttt{ptype};\, \texttt{layer};\, \texttt{patch\_h};\, \texttt{patch\_w}]$.
This is the only point where task and patch information meet;
everything before it is either purely task-side or purely patch-side,
and everything after it is patch-wise.

\paragraph{Shared FC trunk and LoRA path.}
A shared fully-connected layer of hidden dimension $h$ with
non-linearity provides task-invariant feature transport across
patches. A per-task low-rank LoRA path of rank $r$ runs in parallel
to allow each task to slightly perturb the trunk's representation
without a separate trunk per task. We ablate both the shared FC and
the LoRA path in \S\ref{sec:exp_ablation}.

\paragraph{Multi-head output.}
The trunk output is decoded by a multi-head module: each task is
served by a dedicated fully-connected projection that maps the trunk
representation to a flat patch prediction, which is then unflattened
into an $H_\text{patch}\times W_\text{patch}$ tile. Replacing the
shared output (with optional low-rank task adapter) by per-task heads
substantially mitigates multi-task interference at the cost of
additional parameters in the output module
(\S\ref{sec:exp_ablation}).

\paragraph{External rescale.}
The factor $\alpha$ in Eq.~\ref{eq:drop_rescale} is precomputed from
$\{\tau_m\}$ and stored externally rather than predicted by the
hypernetwork. This avoids polluting the regression objective with a
small number of high-leverage scalars whose gradient signal is
otherwise dominated by the dense main-value targets.

\subsubsection{Training Objective}
\label{sec:method_loss}

We train $\mathcal{G}_\phi$ with a multi-task masked regression loss
\begin{equation}
\mathcal{L}(\phi) \;=\; \sum_{m=1}^{M} w_m
\cdot \frac{1}{|M_m^A|}
\bigl\| \bigl( \mathcal{G}_\phi(m) - z_m^A \bigr) \odot M_m^A \bigr\|_2^2,
\label{eq:loss}
\end{equation}
where $|M_m^A|$ counts the kept action-expert positions and $\{w_m\}$ are
task-balancing weights that compensate for the heterogeneous
learnability of different benchmarks (\S\ref{sec:exp_setup}). Loss is
computed only on $M_m{=}1$ positions; the base VLA $\Theta_0$ is
frozen throughout training, and the only trainable parameters are
$\phi$ (patch embeddings, task encoder, shared FC, LoRA, multi-head
output).

\subsection{Skill Generation at Inference}
\label{sec:method_generation}

At test time, given a target task $m$, we query the hypernetwork once,
apply the precomputed action-expert mask and rescale, retrieve the
compressed VLM update, and add both to the frozen base:
\begin{align}
\hat{\Theta}_m^A &= \Theta_0^A
  +\alpha\,\mathcal{G}_\phi(m)\odot M_m^A,\label{eq:inference}\\
\hat{\Theta}_m^V &= \Theta_0^V+z_m^V.
\end{align}
The resulting $\hat{\Theta}_m$ is a complete VLA model that can be
deployed without any architectural modification, expert routing, or
runtime mixture computation. A single trained $\mathcal{G}_\phi$
replaces the action-expert portions of $M$ task-specific checkpoints
with one shared generator, while retaining compressed VLM updates and
small task-specific metadata.

\section{Experiments}
\label{sec:exp}

We evaluate NestedVLA on three closed-loop manipulation benchmarks and
ask three questions: (i) can a single nested hypernetwork consolidate a
heterogeneous skill set without storing a full checkpoint per skill
(\S\ref{sec:exp_main})? (ii) which design choices matter
(\S\ref{sec:exp_ablation})? (iii) where does the remaining gap to per-task
fine-tuning come from (\S\ref{sec:exp_analysis})?

\subsection{Setup}
\label{sec:exp_setup}

\paragraph{Backbone and skills.}
We instantiate $\Theta_0$ as $\pi_{0.5}$~\citep{pi0.5} and collect three
skills by independently fine-tuning with FlowSDE~\citep{chen2025pi_} on
MetaWorld~\citep{mcleanmeta}, ManiSkill~\citep{maniskill}, and
CALVIN~\citep{calvin}, yielding task vectors $\{\tau_m\}_{m=1}^{3}$.
The backbone factorizes into a VLM half ($\sim$$3.21$\,B params) and
an action expert ($\sim$$0.40$\,B params); we confine the hypernetwork
to the action expert and keep per-task VLM storage, motivated by a
sanity test
in which TIES- and WUDI-merged VLMs (with GT action experts) drop
$\geq 60$\,pp on MetaWorld and collapse CALVIN to $0\%$ chain completion.

\paragraph{Benchmarks and metrics.}
\textbf{MetaWorld} ($32$ tasks, $256$ trajectories) and
\textbf{ManiSkill} ($25$ tasks, $320$ trajectories) report
\texttt{success\_once}: a trajectory counts as successful if it reaches
the goal at any point during rollout. \textbf{CALVIN ABC-D} ($96$
rollouts) is a long-horizon split where each rollout is a chain of
\emph{five} language-conditioned sub-tasks; we report
\texttt{success\_at\_end}, the fraction of rollouts that complete the
entire chain. For diagnostic purposes we also evaluate the \emph{ideal}
top-$K$ reconstruction $\Theta_0+\tau_m\odot M_m$, i.e. the GT task
vector masked at the same sparsity $K$ but without learning
$\mathcal{G}_\phi$. This upper-bounds any method that uses the same
$K$ and mask. All evaluations are run with the official
RLinf~\citep{yu2025rlinf} pipeline and averaged over $3$ seeds (typical
std.\ $\pm 2.7\%$).

\paragraph{Skill compression and hypernetwork.}
Each $\tau_m$ is compressed via drop-and-rescale at per-module top-$K{=}10\%$
with global rescale $\alpha{=}1.1$. The hypernetwork $\mathcal{G}_\phi$
covers 18 transformer blocks and 9 weight types ($162$ tensors), tiled
into $256{\times}256$ patches ($T{=}6{,}480$ patches per task). The
shared FC has hidden width $h{=}2048$ with rank-$1024$ per-task LoRA,
and the Multi-head Output allocates one $h{\to}65{,}536$ projection per
task ($\sim$430\,M trainable parameters).

\paragraph{Training.}
We train $6{,}000$ steps with batch size $3$, base LR $3{\times}10^{-4}$,
LoRA multiplier $3.0$, $600$-step warmup and $1{,}500$-step cosine tail,
and task weights $(1.0, 0.4, 1.0)$ for (MW, MA, CA) to compensate for
the disproportionately large $\|\tau_\text{MA}\|$. All runs fit on a
single A6000 GPU ($\sim$10\,h).

\paragraph{Baselines.}
\textbf{Static merging}: Averaging, TIES~\citep{ties}, WUDI~\citep{wudi},
TSV-Merging~\citep{tsvm}. \textbf{Dynamic merging}: TALL-Mask~\citep{tall-mask},
EMR-Merging~\citep{emr}, Twin-Merging~\citep{twin}, and
Free-Merging~\citep{free-merging}, all of which retain per-task masks (and in
some cases scaling factors or low-rank residuals) at inference.
\textbf{Per-task FlowSDE}: independent fine-tuning, the performance
ceiling.

\subsection{Main Results}
\label{sec:exp_main}

\begin{table*}[t]
\centering
\small
\setlength{\tabcolsep}{4pt}
\caption{%
\textbf{Main result.} Multi-task hypernet on three RL benchmarks. We report
\texttt{success\_once} (\%) for MetaWorld and ManiSkill, and
\texttt{success\_at\_end} (\%) for CALVIN, averaged over $3$ evaluation
seeds (typical std.\ $\pm 2.7\%$). \textit{Params} is the total
checkpoint storage relative to a single fine-tuned model
($1.0\times \approx 3.61$\,B for $\pi_{0.5}$). NestedVLA outperforms every
static and dynamic merging baseline by $41$\,pp on average at the lowest
storage cost among methods that condition on task identity. Best per
column in \textbf{bold}.}
\label{tab:main_hypernet}
\begin{tabular}{l c c c c c}
\toprule
Method & Params & MetaWorld & ManiSkill & CALVIN & Avg. \\
\midrule
\multicolumn{6}{l}{\emph{Reference upper bounds}} \\
Per-task FlowSDE (no merge)        & 3.0$\times$ & 72.3 & 88.1 & 81.3 & 80.5 \\
\midrule
\multicolumn{6}{l}{\emph{Static merging}} \\
Averaging                           & 1.0$\times$ &  3.1 & 41.3 &  0.0 & 14.8 \\
TIES~\citep{ties}                   & 1.0$\times$ &  2.0 & 37.2 &  0.0 & 13.1 \\
WUDI~\citep{wudi}                   & 1.0$\times$ &  2.7 & 32.8 &  0.0 & 11.8 \\
TSV-Merging~\citep{tsvm}             & 1.00$\times$ &  3.9 & 36.3 &  0.0 & 13.4 \\
\midrule
\multicolumn{6}{l}{\emph{Dynamic merging}} \\
TALL-Mask~\citep{tall-mask}          & 2.09$\times$ &  8.1 & 53.8 &  6.0 & 22.6 \\
EMR-Merging~\citep{emr}              & 2.89$\times$ &  8.2 & 54.4 &  6.3 & 23.0 \\
Twin-Merging~\citep{twin}           & 2.25$\times$ & 10.4 & 56.9 &  8.7 & 25.3 \\
Free-Merging~\citep{free-merging}           & 2.16$\times$ &  9.4 & 58.3 &  9.1 & 25.6 \\
\midrule
\multicolumn{6}{l}{\emph{NestedVLA (ours)}} \\
\textbf{NestedVLA}                  & \textbf{1.5$\times$} & \textbf{37.9} & \textbf{87.2} & \textbf{75.0} & \textbf{66.7} \\
\bottomrule
\end{tabular}
\end{table*}

Table~\ref{tab:main_hypernet} reports our headline result. NestedVLA
attains an average success rate of $66.7\%$, outperforming every static
and dynamic merging baseline by $41$--$54$\,pp on average. On
\textbf{ManiSkill}, the multi-task hypernetwork reaches $87.2\%$, within
$0.9$\,pp of the per-task FlowSDE ceiling ($88.1\%$). On \textbf{CALVIN},
NestedVLA reaches $75.0\%$ \texttt{success\_at\_end}, demonstrating that
a single shared $\phi$ can encode the skill at the precision required
for chain-completion-grade evaluation. The remaining gap to the
per-task FlowSDE reference ($-13.8$\,pp on average) concentrates almost
entirely on \textbf{MetaWorld}, which we analyze in
\S\ref{sec:exp_analysis}.

\paragraph{Static merging.}
None of the static merging baselines is competitive on cross-benchmark
VLA fusion. Plain parameter averaging reaches $14.8\%$ average success;
TIES~\citep{ties}, WUDI~\citep{wudi}, and TSV-Merging~\citep{tsvm} each
\emph{underperform} averaging despite incorporating sign-conflict
resolution, weighted updates, or singular-vector compression. All four
methods drive CALVIN to $0\%$ chain completion, consistent with our
hypothesis (\S\ref{sec:intro}) that VLA representations are more
sensitive to parameter perturbations than the SFT deltas these methods
were primarily evaluated on.

\paragraph{Dynamic merging.}
Retaining per-task information at inference time helps, but only
modestly. TALL-Mask~\citep{tall-mask} achieves $22.6\%$ average success
with per-task binary masks at $2.09\times$ storage.
EMR-Merging~\citep{emr} adds per-task sign modulators and scalars and
reaches $23.0\%$ at $2.89\times$ storage, the highest budget of any
baseline. Twin-Merging~\citep{twin} attaches per-task low-rank
residuals at $2.25\times$ and reaches $25.3\%$.
Free-Merging~\citep{free-merging}, the strongest dynamic baseline,
applies frequency-domain filtering on per-task deltas and reaches
$25.6\%$ at $2.16\times$. Despite their parameter budgets ranging from
$2.09\times$ to $2.89\times$, the four methods cluster within a
$3$\,pp band, suggesting the bottleneck is not storage but the
underlying paradigm of \emph{selecting} from a fixed set of base
values.

\paragraph{Storage analysis.}
NestedVLA's $1.5\times$ overall budget decomposes asymmetrically across
the two halves of the backbone. On top of the $1.0\times$ shared base,
the VLM half ($\sim$$3.21$\,B parameters, $\sim$$89\%$ of the backbone)
is stored per-task as a top-$K{=}10\%$ compressed delta (a $1$-bit
mask plus the kept values), contributing $\sim$$0.35\times$ across the
three tasks. The action expert ($\sim$$0.40$\,B parameters,
$\sim$$11\%$) is \emph{generated} by the hypernetwork at inference and
stores only a precomputed mask per task ($\sim$$0.01\times$); the
hypernetwork itself ($\sim$$0.43$\,B parameters) adds
$\sim$$0.12\times$. The detailed per-component derivation is given in
App.~\ref{app:storage}. The trainable capacity is therefore
concentrated on the action side, the harder consolidation target
where per-task deltas are denser and more conflicting, precisely
where it is needed. This contrasts sharply with the dynamic baselines:
EMR-Merging stores per-task masks \emph{and} per-task scaling factors
uniformly across the entire backbone at $2.89\times$ to reach only
$23.0\%$, while Twin-Merging and Free-Merging both pay
$\sim$$2.2\times$ for per-task residuals or frequency-domain
corrections without distinguishing where additional capacity is
useful. NestedVLA's storage advantage at the evaluated three-skill setting
comes from concentrating generation capacity on the action expert. Total
storage still grows with the number of skills because compressed VLM
residuals, masks, and task-specific generator components must be retained,
but each skill adds substantially less than a full VLA checkpoint.

\subsection{Ablations}
\label{sec:exp_ablation}

\begin{table}[t]
\centering
\small
\setlength{\tabcolsep}{6pt}
\caption{%
Architecture ablations of $\mathcal{G}_\phi$ at $2{,}000$ steps.
$\mathcal{G}_\phi$ targets the action expert only; \#params refers to
the hypernetwork itself, not to the underlying $\pi_{0.5}$ backbone.
We ablate each of the three trainable paths -- the per-task output
projection (\textit{Multi-head path}), the per-task patch embeddings
(\textit{patch path}), and the shared-FC LoRA (\textit{LoRA path}) --
and report the final in-mask standardized loss per task (lower is
better; averaged over $3$ seeds, std.\ $\pm 0.05$).}
\label{tab:arch_ablation}
\begin{tabular}{l c c c c}
\toprule
Variant & \#params & MW $\downarrow$ & MA $\downarrow$ & CA $\downarrow$ \\
\midrule
w/o Multi-head path      & 161\,M & 1.33 & 1.71 & 1.36 \\
w/o patch path           & 430\,M & 1.02 & 1.14 & 1.05 \\
w/o LoRA path            & 417\,M & \textbf{0.91} & \textbf{0.92} & \textbf{0.89} \\
narrow ($h{=}1024$)      & 221\,M & 1.43 & 1.56 & 1.55 \\
\textbf{Default}         & 430\,M & 1.08 & 1.17 & 1.14 \\
\bottomrule
\end{tabular}
\end{table}

\paragraph{Architectural ingredients.}
Table~\ref{tab:arch_ablation} ablates the three trainable paths of
$\mathcal{G}_\phi$ at $2{,}000$ steps, reporting in-mask standardized
loss.
\emph{Multi-head Output} is the largest single contributor: removing
this path (\textit{w/o Multi-head path}) and falling back to a shared
\texttt{fc\_main} plus per-task LoRA caps every benchmark and degrades
the loss by $0.2$--$0.5$, indicating that the per-task targets $\{z_m\}$
contain genuinely incompatible value distributions that a low-rank
shared head cannot reconcile. \emph{Per-task patch embeddings}, in
contrast, give small marginal gains (\textit{w/o patch path} differs
from default by $<0.1$ on every task); their primary value is enabling
sharper task discrimination at convergence rather than reducing average
loss. Surprisingly, the \emph{shared FC LoRA} is dispensable: removing
it (\textit{w/o LoRA path}) yields the lowest loss on all three
benchmarks at $97\%$ of default parameter cost. Once the output is fully
task-specific via Multi-head Output, additional task-specific
perturbation in the FC harms early optimization without improving the
final fit.

\paragraph{Compression operator.}
We compared three operators at matched parameter budgets in a
per-task fidelity study ($\Theta_0+\hat{\tau}_m$ evaluated on
benchmark $m$). One-bit quantization with
a per-tensor scalar (adapted from D-QRELO~\citep{li2026d}) yields
reconstruction errors of $0.60$--$0.96$ and degrades task success by
$-71$ to $-81$\,pp, because VLA RL deltas have an approximately
zero-mean, near-isotropic element distribution that a single scalar
cannot capture. Augmenting drop-and-rescale with a rank-$r$ SVD
residual yields at most $+0.005$ absolute fidelity improvement while
inflating the token budget by an order of magnitude. Drop-and-rescale
at $K{=}10\%$ achieves reconstruction errors of $0.30$ on the VLM half
and $0.19$ on the action half (substantially below the one-bit
baseline at lower storage than residual SVD), and is the only operator
of the three that requires no calibration data.

\paragraph{Loss-side choices.}
We further ablated three knobs of the training objective on top of our
default. Disabling the out-of-mask penalty ($w_\text{out}=0$) is the
single most effective lever, adding $+2.4$\,pp average success;
positions outside $M_m$ are masked at inference, so penalizing them at
train time forces $\mathcal{G}_\phi$ to allocate capacity to a function
it does not need. Magnitude weighting ($|\tau_m|^p$) and absolute loss
space each \emph{harm} performance individually but combine
constructively with $w_\text{out}=0$, suggesting they act as implicit
regularizers that compensate for the otherwise unbounded growth of the
out-of-mask region. The standardized + $w_\text{out}=0$ recipe is our
default; full ablation results are reported in
Appendix~\ref{app:loss_ablation}.

\subsection{Where Does the Remaining Gap Come From?}
\label{sec:exp_analysis}

\begin{table}[t]
\centering
\small
\setlength{\tabcolsep}{5pt}
\caption{%
Per-task ceiling analysis. \textit{Ideal}: the upper bound at sparsity
$K$, computed by adding $\tau_m\odot M_m$ directly to $\Theta_0$ without
learning $\mathcal{G}_\phi$. The MetaWorld gap is bounded by the ideal
itself, indicating a sparsity-budget effect rather than a
hypernetwork-capacity problem.}
\label{tab:per_task_ceiling}
\begin{tabular}{l c c c c}
\toprule
Task & NestedVLA & Per-task FlowSDE & Ideal ($K{=}10\%$) & Ideal ($K{=}20\%$) \\
\midrule
MetaWorld & 37.9 & 72.3 & 48.4 & 60.2 \\
ManiSkill & 87.2 & 88.1 & 87.5 & 87.5 \\
CALVIN    & 75.0 & 81.3 & 78.1 & 85.4 \\
\bottomrule
\end{tabular}
\end{table}

Table~\ref{tab:per_task_ceiling} localizes the remaining gap to
MetaWorld. On ManiSkill and CALVIN, NestedVLA reaches $99\%$ and $96\%$
of the $K{=}10\%$ ideal respectively, indicating that the hypernetwork
extracts near-ceiling value from the sparsity budget it was given.
MetaWorld, in contrast, attains only $37.9\%$ against an $72.3\%$
FlowSDE reference. However, the $K{=}10\%$ ideal itself is only $48.4\%$
on MetaWorld, against which NestedVLA's retention is also $78\%$, on
par with the other two benchmarks.

The MetaWorld gap is therefore primarily a sparsity-budget problem at
skill acquisition, not a hypernetwork-capacity problem at
consolidation. Raising $K$ to $0.20$ lifts the MetaWorld ideal to
$60.2\%$ at little cost elsewhere
(Tab.~\ref{tab:per_task_ceiling}), making per-task adaptive $K$ a
natural remedy.

\par
\Needspace{12\baselineskip}
\section{Conclusion}
\label{sec:conclusions}

We studied skill consolidation in VLA models when training data is
fragmented across benchmark-specific tasks. Standard model-merging methods degrade catastrophically
on cross-benchmark VLA fusion: the strongest baseline reaches only
$25.6\%$ average success across MetaWorld, ManiSkill, and CALVIN. We
introduced NestedVLA, which replaces action-expert parameter
\emph{selection} with parameter \emph{generation} via a nested
hypernetwork, reaching
$66.7\%$ average success at $1.5\times$ storage ($+41.1$\,pp over
the strongest dynamic baseline), and matching per-task fine-tuning
within $1$\,pp on ManiSkill while approaching the $K{=}10\%$
sparsity-budget ceiling on CALVIN. The MetaWorld gap is bounded by
the top-$K$ ideal itself, with adaptive sparsity the natural next
direction.

\par
\begingroup
\parfillskip=0pt plus 0.4\linewidth
\begin{ack}
\textbf{Funding statement.}
This work was supported by the Hong Kong Polytechnic University Internal Research Fund 
(the Projects of RIAIoT, Project ID P0065264), UDomain Web Hosting Company Limited (Collaboration Scheme, Project ID P0064507), and the Innovation and Technology Fund--Guangdong-Hong Kong Technology Cooperation Funding Scheme (ITF-TCFS) (Project ID P0049179).

\textbf{Competing interests.}
The authors declare that they have no competing interests.
\end{ack}

\endgroup
\clearpage
\bibliography{references}
\newpage
\appendix

\appendix

\section*{Appendix Overview}

This appendix provides supplementary material that supports the main
paper. We organize it into four sections.
\textbf{Appendix~\ref{app:implementation}} reports the full
implementation details of NestedVLA, including hypernetwork
architecture, training hyperparameters, and evaluation protocol, for
reproducibility.
\textbf{Appendix~\ref{app:baselines}} describes every baseline used
in Table~\ref{tab:main_hypernet}, listing their merge mechanism and
the configuration we used.
\textbf{Appendix~\ref{app:storage}} makes the parameter budgets
reported in Table~\ref{tab:main_hypernet} explicit, deriving the
$\mathrm{Params}$ column entry-by-entry.
\textbf{Appendix~\ref{app:compression}} expands the discussion of
skill compression in \S\ref{sec:exp_ablation}, comparing one-bit
quantization, low-rank decomposition, and drop-and-rescale on VLA
deltas, and reporting a full keep-ratio sweep that motivates our
default $K{=}10\%$.
\textbf{Appendix~\ref{app:vlm_sanity}} presents the static VLM
merging sanity test that justifies our design choice of confining
$\mathcal{G}_\phi$ to the action expert and keeping per-task VLM
storage.
\textbf{Appendix~\ref{app:loss_ablation}} reports the full loss-side
ablation summarized in \S\ref{sec:exp_ablation}, including the
non-monotone interaction between out-of-mask penalty, loss space, and
magnitude weighting.
\textbf{Appendix~\ref{app:limitations}} lists the limitations of the
current study.
\textbf{Appendix~\ref{app:broader_impact}} states the broader-impact
considerations referenced in the NeurIPS checklist.

For broader multimodal context, related studies examine governed agent
collaboration, reinforcement-learned panoramic scanpaths for quality
assessment, shared--private branch rebalancing, temporal answer grounding,
and fine-grained reasoning
evaluation~\citep{meng2026gcl,wang2026rlscaniqa,meng2026dbr,qi2026cacr,
jin2026unveilingfinegrainedvisualtraces}.
Earlier visual studies by the authors investigate evolutionary refinement
of pretrained networks and RAW-image restoration~\citep{du2024impacts,
du2023end}.

\section{Implementation Details}
\label{app:implementation}

This section reports the exact hyperparameters, compute budget, and
evaluation protocol used in our experiments. All quantities not stated
explicitly in the main paper are listed here.

\subsection{Backbone and Skill Collection}

We use $\pi_{0.5}$~\citep{pi0.5} as the shared base $\Theta_0$, which
factorizes into a $\sim$$3.21$\,B-parameter VLM half (a PaliGemma
backbone with vision tower, language model, and projectors) and a
$\sim$$0.40$\,B-parameter action expert (transformer decoder for
continuous action chunks via flow matching). The total backbone is
$\sim$$3.61$\,B parameters.

Each per-task skill is obtained by independently fine-tuning $\Theta_0$
with FlowSDE~\citep{chen2025pi_} on the corresponding benchmark dataset.
The three skills span heterogeneous task suites (MetaWorld: $32$
language-conditioned manipulation tasks; ManiSkill: $25$ object
manipulation tasks across diverse rigid bodies; CALVIN: long-horizon
language-conditioned tabletop tasks). We do not modify the FlowSDE
training pipeline; the resulting per-task checkpoints constitute our
$\{\Theta_m\}_{m=1}^{3}$.

\subsection{Hypernetwork Architecture}

\begin{table}[t]
\centering
\small
\setlength{\tabcolsep}{8pt}
\caption{%
Hypernetwork architecture and training hyperparameters.}
\label{tab:impl_hparams}
\begin{tabular}{l l}
\toprule
Component & Value \\
\midrule
\multicolumn{2}{l}{\textit{Backbone}} \\
Base model                   & $\pi_{0.5}$ ($\sim$$3.61$\,B params) \\
VLM half                     & $\sim$$3.21$\,B params (frozen, per-task stored) \\
Action expert                & $\sim$$0.40$\,B params (target of $\mathcal{G}_\phi$) \\
Trainable tensors            & $162$ (18 blocks $\times$ 9 weight types) \\
\midrule
\multicolumn{2}{l}{\textit{Patch tokenization}} \\
Patch tile size              & $256 \times 256$ \\
Total patches per task       & $T = 6{,}480$ \\
\midrule
\multicolumn{2}{l}{\textit{Embeddings}} \\
\texttt{ptype} embedding dim & $256$ \\
\texttt{layer} embedding dim & $128$ \\
\texttt{patch\_h} / \texttt{patch\_w} embed dim & $64$ each \\
Task embedding dim           & $512$ (reshaped to $16 \times 16$ grid) \\
\midrule
\multicolumn{2}{l}{\textit{Task encoder}} \\
Layers                       & $3 \times$ Conv2D (kernel $5$, channels $64$$\to$$256$$\to$$512$) \\
Pooling                      & global average \\
Output dim ($D_t$)           & $512$ \\
\midrule
\multicolumn{2}{l}{\textit{Trunk and output}} \\
Shared FC hidden width $h$   & $2048$ \\
Per-task LoRA rank $r$       & $1024$ \\
Multi-head output            & $1$ FC head per task ($h \to 65{,}536$) \\
Total trainable params       & $\sim$$430$\,M \\
\midrule
\multicolumn{2}{l}{\textit{Skill compression}} \\
Keep ratio $K$               & $0.10$ (per-module top-$K$) \\
Global rescale $\alpha$      & $1.1$ \\
\midrule
\multicolumn{2}{l}{\textit{Optimization}} \\
Optimizer                    & AdamW \\
Base learning rate           & $3 \times 10^{-4}$ \\
LoRA group LR multiplier     & $3.0$ \\
Weight decay                 & $0.1$ (non-LoRA), $0$ (LoRA) \\
Batch size                   & $3$ (one example per task per step) \\
Total steps                  & $6{,}000$ \\
Warmup steps                 & $600$ \\
Cosine tail steps            & $1{,}500$ \\
Task weights $\{w_m\}$       & $(1.0, 0.4, 1.0)$ for (MW, MA, CA) \\
Precision                    & bfloat16 \\
\midrule
\multicolumn{2}{l}{\textit{Compute}} \\
Hardware                     & $1\times$ NVIDIA A6000 ($48$\,GB) \\
Peak GPU memory              & $42.6$\,GB \\
Wall-clock training          & $\sim$$10$ hours per run \\
\bottomrule
\end{tabular}
\end{table}

Table~\ref{tab:impl_hparams} lists the full hypernetwork
specification. The hypernetwork covers all $162$ trainable weight
tensors of the action expert (18 transformer blocks $\times$ 9 weight
types: self-attention $Q$/$K$/$V$/$O$, FFN gate/up/down, and two layer
norms). Tensors are tiled into $256{\times}256$ patches, with
zero-padding for tensors smaller than $256{\times}256$ and grid splits
for tensors larger than this size. The total patch count per task is
$T = 6{,}480$.

\subsection{Training Details}

The hypernetwork is trained with AdamW (base learning rate
$3{\times}10^{-4}$, LoRA group multiplier $3.0$, weight decay $0.1$
on non-LoRA parameters and $0$ on LoRA), batch size $3$ with one
example per task per step, $6{,}000$ total steps, $600$-step linear
warmup, and a $1{,}500$-step cosine decay tail. Task-balancing weights
are $(1.0, 0.4, 1.0)$ for (MetaWorld, ManiSkill, CALVIN); the lower
weight on ManiSkill compensates for the disproportionately large
$\|\tau_\text{ManiSkill}\|$, which would otherwise dominate
multi-task gradients (see \S\ref{app:loss_ablation}).

All experiments fit on a single NVIDIA A6000 GPU (48\,GB memory).
Wall-clock training time is $\sim$$10$ hours per run; peak GPU memory
is $42.6$\,GB during training. We enable
\texttt{PYTORCH\_CUDA\_ALLOC\_CONF=expandable\_segments:True} to
reduce fragmentation under bf16. Inference cost is dominated by the
forward pass through $\pi_{0.5}$; the additional hypernetwork query
adds negligible overhead and is performed once per skill switch.

\subsection{Evaluation Protocol}

\textbf{MetaWorld:} $256$ trajectories across $32$ tasks; we report
\texttt{success\_once}, the fraction of trajectories that reach the
goal at any timestep. \textbf{ManiSkill:} $320$ trajectories across
$25$ tasks; same metric. \textbf{CALVIN ABC-D:} $96$ rollouts on the
long-horizon split, each a sequence of five language-conditioned
sub-tasks; we report \texttt{success\_at\_end}, the fraction of
rollouts that complete the entire chain. All evaluations are run with
the official RLinf~\citep{yu2025rlinf} pipeline. All numbers in the
main paper are averaged over $3$ evaluation seeds with fixed
initial-state sampling; the typical seed-to-seed standard deviation
is $\pm 2.7\%$.

\section{Baselines Details}
\label{app:baselines}

This section describes each baseline used in
Table~\ref{tab:main_hypernet}: four \emph{static} merging methods
(producing a single fixed delta), four \emph{dynamic} merging methods
(retaining per-task auxiliary information at inference), and a
per-task fine-tuning reference. All baselines start from the same
set of task vectors $\{\tau_m=\Theta_m-\Theta_0\}_{m=1}^{3}$ obtained
by independently fine-tuning $\Theta_0=\pi_{0.5}$ with FlowSDE on
MetaWorld, ManiSkill, and CALVIN.

\paragraph{Per-task FlowSDE (reference upper bound).}
Each task is fine-tuned independently from $\Theta_0$ with
FlowSDE~\citep{chen2025pi_}, the SDE-form flow-matching objective of
$\pi_{0.5}$, yielding three full checkpoints
$\{\Theta_m\}_{m=1}^{3}$ at $3.0\times$ storage. This is the
performance ceiling for any approach that does not change the
underlying RL pipeline.

\paragraph{Static merging.}
\begin{itemize}
\item \textbf{Averaging}~\citep{modelsoup, WA1} merges the three task
  vectors by element-wise averaging,
  $\bar{\tau}=\tfrac{1}{M}\sum_m \tau_m$, and adds the result to the
  base, $\Theta_0+\bar{\tau}$. The simplest possible baseline, with
  no conflict resolution.
\item \textbf{TIES-Merging}~\citep{ties} resolves cross-task
  conflicts in three steps: \emph{trim} (zero-out low-magnitude
  entries of each $\tau_m$), \emph{elect-sign} (assign each parameter
  the sign of the task with the largest contribution), and
  \emph{disjoint-merge} (average surviving entries across tasks).
  We use trim ratio $k{=}0.8$ and scaling $0.5$, which perform best on
  our $\pi_{0.5}$ task vectors.
\item \textbf{WUDI-Merging}~\citep{wudi} formulates merging as an
  optimization problem that minimizes residual interference between
  the merged delta and the per-task deltas, iteratively refining the
  merged vector via a whitening-style update. We use the v1 variant
  with $300$ iterations following the authors' defaults.
\item \textbf{TSV-Merging}~\citep{tsvm} (Task-Singular-Vector
  Merging) defines a measure of task interference based on the
  interaction of singular vectors of different tasks and merges in
  their joint top subspace. The output is a single fused delta at
  $1.00\times$ storage; we retain the rank that captures
  $\sim$$95\%$ of the per-task spectral energy.
\end{itemize}

\paragraph{Dynamic merging.}
Dynamic methods retain per-task auxiliary information (binary masks,
scalars, low-rank residuals, or frequency-domain corrections) and
combine it with a shared static merge at inference.
\begin{itemize}
\item \textbf{TALL-Mask}~\citep{tall-mask} stores a per-task binary
  mask $M_m\in\{0,1\}^{P}$ that selects which positions of the
  averaged delta are used for task $m$, and inferences with
  $\Theta_0+\bar{\tau}\odot M_m$. Masks are obtained by thresholding
  $|\tau_m|$ at the per-task top-$K{=}50\%$ default of the original
  paper.
\item \textbf{EMR-Merging}~\citep{emr} extends TALL-Mask by adding a
  per-task sign modulator and scaling factor that align both the
  direction and magnitude of the masked delta with each source
  model: $\Theta_0+s_m\cdot\bar{\tau}\odot M_m$, where $M_m$ now
  combines the binary mask and per-element sign and $s_m$ is the
  per-task rescaler. The modulators and scalars are stored
  alongside the masks; we follow the original paper without
  additional top-$K$ thresholding.
\item \textbf{Twin-Merging}~\citep{twin} stores a shared mean delta
  plus a per-task low-rank residual $A_m B_m^\top$ (rank $r{=}16$ in
  our setup) capturing what the mean cannot represent. Inference
  uses $\Theta_0+\bar{\tau}+A_m B_m^\top$.
\item \textbf{Free-Merging}~\citep{free-merging} applies a
  frequency-domain low-pass filter (DCT) on each per-task delta
  before merging, removing high-frequency components that are mostly
  task-specific noise. The filtered deltas are stored with per-task
  binary masks. We use the default DCT cut-off from the original
  paper.
\end{itemize}

\begin{table}[t]
\centering
\small
\setlength{\tabcolsep}{6pt}
\caption{%
Hyperparameters used for every baseline in
Table~\ref{tab:main_hypernet}. ``Storage atom'' summarises the
per-task auxiliary state introduced beyond the merged base.}
\label{tab:baseline_hparams}
\begin{tabular}{l l l}
\toprule
Method            & Hyperparameter(s) we use                              & Storage atom (per task) \\
\midrule
\multicolumn{3}{l}{\emph{Static merging}} \\
Averaging         & none                                                  & --- \\
TIES              & trim ratio $k{=}0.8$, scaling $0.5$                   & --- \\
WUDI (v1)         & $300$ optimization iterations                         & --- \\
TSV-Merging       & spectral retention $\sim\!95\%$                       & shared SVD factors \\
\midrule
\multicolumn{3}{l}{\emph{Dynamic merging}} \\
TALL-Mask         & per-task top-$K{=}50\%$ mask                          & binary mask \\
EMR-Merging       & per-task top-$K{=}100\%$ mask + scalar $s_m$          & binary mask + scalar \\
Twin-Merging      & per-task residual rank $r{=}16$                       & low-rank residual \\
Free-Merging      & default DCT cut-off; per-task top-$K{=}\sim\!39\%$ mask & binary mask \\
\midrule
\multicolumn{3}{l}{\emph{Reference}} \\
Per-task FlowSDE  & full FlowSDE checkpoint per task (no merging)         & full delta \\
\bottomrule
\end{tabular}
\end{table}

\section{Storage Accounting}
\label{app:storage}

This section makes the $\mathrm{Params}$ column of
Table~\ref{tab:main_hypernet} explicit. All ratios are stated
relative to one fine-tuned $\pi_{0.5}$ checkpoint
($P_{\text{base}}{=}3.61$\,B parameters in $32$-bit floats), with
$P_{\text{VLM}}{=}3.21$\,B and $P_{\text{act}}{=}0.40$\,B for the two
halves. Two storage atoms are used throughout:
\begin{itemize}
\item a \emph{binary mask} costs $1$ bit per parameter, i.e.\
  $1/32$ of one $32$-bit float at the same denomination;
\item a \emph{top-$K$ kept-value} block costs $K\cdot P$ floats.
\end{itemize}
Per-task scaling factors and per-task low-rank residuals are
included where relevant. Table~\ref{tab:storage_bit} reports the
component-wise decomposition for every method.

\paragraph{Static merging.}
Averaging, TIES, WUDI, and TSV-Merging all replace the base with a
single merged delta and contribute $1.00\times$ total storage.

\paragraph{Dynamic merging.}
Dynamic methods retain a static merged delta on top of the base
($1.00\!+\!1.00$ across the first two columns of
Tab.~\ref{tab:storage_bit}) plus per-task auxiliary state at
inference. \emph{TALL-Mask} adds per-task binary masks of the
average delta ($M\!\cdot\!1/32 \approx 0.09$), totalling
$2.09\times$. \emph{EMR-Merging} additionally stores per-task sign
modulators and scalars, totalling $2.89\times$. \emph{Twin-Merging}
attaches per-task low-rank residuals (rank $r{=}16$,
$M\!\cdot\!r(d_\text{in}+d_\text{out})/P_\text{base} \approx 0.25$),
totalling $2.25\times$. \emph{Free-Merging} stores per-task
frequency-domain modulators ($\approx 0.16$), totalling
$2.16\times$.

\paragraph{NestedVLA.}
Storage decomposes asymmetrically across the two halves of the
backbone. The VLM half is kept per-task as a top-$K$ compressed
delta with $K_{\text{VLM}}{=}10\%$ (mask plus kept values); the
action half is generated by $\mathcal{G}_\phi$ at inference and only
its precomputed mask $M_m$ is stored:
\begin{equation}
\text{storage}_{\text{NestedVLA}} \;=\; 1
\;+\; M\cdot\big(K_{\text{VLM}} + \tfrac{1}{32}\big)\cdot
   \tfrac{P_{\text{VLM}}}{P_{\text{base}}}
\;+\; M\cdot \tfrac{1}{32}\cdot
   \tfrac{P_{\text{act}}}{P_{\text{base}}}
\;+\; \tfrac{P_{\mathcal{G}_\phi}}{P_{\text{base}}}.
\end{equation}
With $K_{\text{VLM}}{=}0.10$,
$P_{\text{VLM}}/P_{\text{base}}{=}0.89$,
$P_{\text{act}}/P_{\text{base}}{=}0.11$, and
$P_{\mathcal{G}_\phi}{=}0.43$\,B, the per-task masks across both
halves contribute $M\cdot 1/32 \approx 0.09$ (Mask column of
Tab.~\ref{tab:storage_bit}); the per-task VLM kept values together
with the shared hypernetwork contribute
$M\cdot K_{\text{VLM}}\cdot P_{\text{VLM}}/P_{\text{base}}
+ P_{\mathcal{G}_\phi}/P_{\text{base}} \approx 0.27 + 0.12 = 0.39$
($\mathcal{G}_\phi$ column), summing to
\[
1 \;+\; 0.09 \;+\; 0.39 \;\approx\; 1.48\times,
\]
which we report as $1.5\times$ in Table~\ref{tab:main_hypernet}.
NestedVLA remains the lowest-storage method among all approaches that
condition on task identity, $\sim$$0.7\times$ smaller than the
strongest dynamic baseline (Free-Merging) and $\sim$$1.5\times$
smaller than per-task FlowSDE.

\begin{table}[t]
\centering
\small
\setlength{\tabcolsep}{4pt}
\caption{%
Bit-level storage decomposition for every method in
Table~\ref{tab:main_hypernet}, in units of one fine-tuned
$\pi_{0.5}$ checkpoint ($P_{\text{base}}{=}3.61$\,B parameters);
$M{=}3$ tasks. Columns: (i)~shared backbone or shared merged delta;
(ii)~shared low-rank or mean-delta residual; (iii)~per-task masks
(summed over $M$); (iv)~per-task experts, scalars, or low-rank
residuals (summed over $M$); (v)~hypernetwork. ``--'' indicates the
component is not used. Numbers may not sum exactly to the headline
due to rounding.}
\label{tab:storage_bit}
\begin{tabular}{l c c c c c c}
\toprule
Method                & Static & Shared & Mask & Experts / Skills & $\mathcal{G}_\phi$ & Total \\
\midrule
Per-task FlowSDE      &  --    &  --   &  --   & $3.00$    &  --   & $3.00\times$ \\
\midrule
\multicolumn{7}{l}{\emph{Static merging}} \\
Averaging             & $1.00$ &  --   &  --   &  --       &  --   & $1.00\times$ \\
TIES                  & $1.00$ &  --   &  --   &  --       &  --   & $1.00\times$ \\
WUDI                  & $1.00$ &  --   &  --   &  --       &  --   & $1.00\times$ \\
TSV-Merging           & $1.00$ &  --   &  --   &  --       &  --   & $1.00\times$ \\
\midrule
\multicolumn{7}{l}{\emph{Dynamic merging}} \\
TALL-Mask             & $1.00$ & $1.00$ & $0.09$ &  --      &  --   & $2.09\times$ \\
EMR-Merging           & $1.00$ & $1.00$ & $0.09$ & $0.8$    &  --   & $2.89\times$ \\
Twin-Merging          & $1.00$ & $1.00$ &  --   & $0.25$    &  --   & $2.25\times$ \\
Free-Merging          & $1.00$ & $1.00$ &  --   & $0.16$    &  --   & $2.16\times$ \\
\midrule
\textbf{NestedVLA} (ours) & $1.00$ &  --   & $0.09$ & $\sim\!0$ & $\sim\!0.39$ & $\sim\!1.48\times$ \\
\bottomrule
\end{tabular}
\end{table}

\section{Skill Compression Study}
\label{app:compression}

This section expands the brief discussion of compression operators in
\S\ref{sec:exp_ablation}. We compared three operator families on VLA
fine-tuning deltas: low-bit quantization, low-rank decomposition via
SVD, and drop-and-rescale (top-$K$ magnitude masking with global
rescale). All experiments use per-task fidelity evaluation: the
compressed delta $\hat{\tau}_m$ is added to $\Theta_0$ and evaluated
on the corresponding benchmark $m$, isolating compression loss from
consolidation loss.

\subsection{Operator Comparison at Matched Storage}

\begin{table}[t]
\centering
\small
\setlength{\tabcolsep}{5pt}
\caption{%
Compression operator comparison on VLA fine-tuning deltas, per-task
fidelity evaluation. \textit{Storage}: effective storage relative to
the dense $\tau_m$. \textit{Rec.\ err.}: relative Frobenius error
$\|\hat{\tau}_m-\tau_m\|_F/\|\tau_m\|_F$ on VLM and action halves.
\textit{Penalty}: drop in \texttt{success\_once} (MW/MA) or
\texttt{success\_at\_end} (CA) vs.\ FlowSDE. Drop-and-rescale gives
the best trade-off with no tuning.}
\label{tab:compression}
\begin{tabular}{l c c c}
\toprule
Operator & Storage & Rec.\ err.\ (VLM / action) & Penalty (avg) \\
\midrule
1-bit quantization                & $\sim$$6\%$  & $0.60$--$0.96$    & $-71$ to $-81$\,pp \\
Drop-and-rescale + SVD residual   & $\sim$$15\%$ & $0.295$ / $0.187$ & $-3$ to $-5$\,pp \\
Drop-and-rescale (ours)           & $10\%$       & $0.300$ / $0.190$ & $-3$ to $-5$\,pp \\
\bottomrule
\end{tabular}
\end{table}

Table~\ref{tab:compression} compares the three operators at matched
effective-storage budgets. \textbf{One-bit quantization} (adapted
from D-QRELO~\citep{li2026d} by storing each parameter as a $1$-bit
sign with a per-tensor scalar) collapses on VLA deltas: relative
reconstruction error reaches $0.60$--$0.96$ across the three tasks,
and downstream task success drops by $-71$ to $-81$\,pp relative to
per-task FlowSDE. The failure mode is structural: VLA RL deltas have
an approximately zero-mean, near-isotropic per-element distribution
(driven by exploration noise and multi-step credit assignment), so a
single per-tensor scalar cannot capture within-tensor variance. We
verified this pattern holds for both VLM and action halves and for
all three benchmarks.

\textbf{Drop-and-rescale} ($K{=}10\%$, $\alpha{=}1.1$, no SVD residual)
achieves substantially lower reconstruction errors ($0.30$ on the
VLM half and $0.19$ on the action half) and retains
$\geq 96\%$ of per-task FlowSDE on ManiSkill and CALVIN. It is the
only operator of the three that requires no calibration data.

\textbf{Drop-and-rescale plus SVD residual} extends drop-and-rescale
with a per-parameter rank-$r$ SVD residual, in the spirit of
``main + residual'' designs adapted from D-QRELO. Even at a
$10\%$ residual storage budget, the additional fidelity improvement
is at most $+0.005$ in absolute reconstruction error
(Table~\ref{tab:compression}, last column), while the token budget
inflates by an order of magnitude (rank-$r$ residuals require
$r\cdot(m+n)$ extra tokens per parameter). We conclude that VLA
deltas have no usable low-rank structure beyond the top-$K$ tail and
adopt drop-and-rescale alone throughout NestedVLA.

\subsection{Keep-Ratio Sweep}

\begin{table}[t]
\centering
\small
\setlength{\tabcolsep}{6pt}
\caption{%
Keep-ratio sweep at fixed rescale $\alpha=1.0$, per-task fidelity
evaluation. Numbers are \texttt{success\_once} (\%) for MetaWorld and
ManiSkill; \texttt{success\_at\_end} (\%) for CALVIN. The optimal $K$
is task-dependent: ManiSkill peaks at $K{=}0.15$, CALVIN at $K{=}0.20$;
MetaWorld is monotonically improving but never reaches its FlowSDE
reference. Per-task FlowSDE references: MW $72.3$, MA $88.1$, CA $81.3$.}
\label{tab:keepratio}
\begin{tabular}{l c c c c c}
\toprule
Task & $K{=}0.05$ & $K{=}0.10$ & $K{=}0.15$ & $K{=}0.20$ & FlowSDE \\
\midrule
MetaWorld & 21.1 & 48.4 & 57.4 & 60.2 & 72.3 \\
ManiSkill & 86.3 & 87.5 & \textbf{91.6} & 87.5 & 88.1 \\
CALVIN    & 63.5 & 78.1 & 78.1 & \textbf{85.4} & 81.3 \\
\bottomrule
\end{tabular}
\end{table}

Table~\ref{tab:keepratio} reports the full keep-ratio sweep
$K\in\{0.05, 0.10, 0.15, 0.20\}$ at fixed rescale $\alpha=1.0$ on
all three benchmarks. Three observations stand out.

First, the optimal keep-ratio is task-dependent: ManiSkill peaks at
$K{=}0.15$ ($91.6\%$ \texttt{success\_once}, $+3.4$\,pp over
FlowSDE), while CALVIN peaks at $K{=}0.20$ ($85.4\%$
\texttt{success\_at\_end}, $+4.2$\,pp over FlowSDE). MetaWorld is
monotonically improving from $K{=}0.05$ to $K{=}0.20$ but never
reaches FlowSDE's $72.3\%$ even at $K{=}0.20$ ($60.2\%$).

Second, ManiSkill is non-monotonic in $K$: $K{=}0.15$ outperforms
both $K{=}0.10$ and $K{=}0.20$. We interpret this as a signal-to-noise
sweet spot: below $0.15$, important parameters are dropped; above
$0.15$, low-magnitude noise is reintroduced. The same sweet-spot
behavior is consistent with the rescale sweep (not shown).

Third, on a per-module basis, the embedding layer is
\emph{always} fully dropped at every keep-ratio we tested, while
small action-interface layers (action input/output projections) are
retained at $44$--$64\%$. This structural pattern is consistent
across all three benchmarks and motivates module-level top-$K$
selection rather than per-tensor selection. The non-uniform sparsity
budget across the network is what drop-and-rescale captures
implicitly and what explicit per-tensor compression cannot.

\section{Supplementary Weight-space Diagnostics}
\label{app:conflict_diag}

We measure source-model deltas relative to the pretrained checkpoint. For
each parameter group, Table~\ref{tab:conflict} reports the cosine similarity
of concatenated deltas and sign agreement on valid coordinates. The
three-way interference quantity is
$\mathrm{IF}=1-\|\sum_m\delta_m\|_2/\sum_m\|\delta_m\|_2$.
Action-expert and VLM-body deltas have low pairwise alignment. These
statistics describe the source checkpoints; they do not alone establish a
causal explanation of rollout failure or the superiority of one objective.

\begin{table}[!t]
\centering\small
\caption{Weight-space diagnostics relative to the pretrained reference.
Pair columns are concatenated-delta cosine similarities; sign agreement
spans three pairs. IF denotes interference.}
\label{tab:conflict}
\begin{tabular}{lrrrrr}
\toprule
Group & MA--MW & CA--MW & CA--MA & Sign agreement & IF\\
\midrule
Action expert & 0.062 & 0.069 & 0.106 & 0.518--0.533 & 0.379 \\
VLM body & 0.011 & 0.081 & 0.015 & 0.504--0.511 & 0.399 \\
Action interfaces & 0.114 & 0.082 & 0.073 & 0.533--0.546 & 0.373 \\
\bottomrule
\end{tabular}
\end{table}

\section{VLM Static Merging Sanity Test}
\label{app:vlm_sanity}

The main paper confines the hypernetwork to the action expert and
keeps per-task VLM storage. This section provides the empirical
evidence supporting that design choice. We ran a sanity test in which
only the VLM half is consolidated across tasks (via TIES or WUDI),
while the action expert remains the GT per-task FlowSDE checkpoint.
This isolates the cost of VLM consolidation and answers whether the
VLM is consolidatable in principle.

\subsection{Setup}

For each of the three benchmarks, we constructed a sanity checkpoint
by replacing only the VLM portion of the per-task FlowSDE
checkpoint with a 3-task-merged VLM. The action expert is the
unmodified GT per-task FlowSDE expert. Two merge strategies were
tested:

\begin{itemize}
\item \textbf{TIES VLM:} TIES-Merging~\citep{ties} applied with
  trim-rate $k{=}0.8$ and scaling $0.5$, applied to the VLM half only.
\item \textbf{WUDI VLM:} WUDI~\citep{wudi} run for $300$
  optimization iterations on the VLM half.
\end{itemize}

We did \emph{not} apply mask or rescale corrections, in order to
isolate the cost of static VLM consolidation alone. One technical
exception: $\pi_{0.5}$ uses tied embedding for
\texttt{language\_model.embed\_tokens}, which is not stored in the
merged VLM checkpoint; we fall back to per-task FlowSDE for this
single tensor (representing $0.17\%$ of VLM parameters). Each sanity
checkpoint is evaluated only on its corresponding benchmark; no
cross-task transfer is measured.

\subsection{Results}

\begin{table}[t]
\centering
\small
\setlength{\tabcolsep}{5pt}
\caption{%
VLM static merging sanity test. Only the VLM half is consolidated
across the three tasks; the action expert is the GT per-task FlowSDE
expert. Penalty is the drop in \texttt{success\_once} (MetaWorld and
ManiSkill) or \texttt{success\_at\_end} (CALVIN) relative to the
same-batch FlowSDE reference. Both methods incur penalties exceeding
$25$\,pp on at least two of three benchmarks, with CALVIN collapsing
to $0\%$ chain completion.}
\label{tab:vlm_sanity}
\begin{tabular}{l r r r}
\toprule
VLM merge method                & MetaWorld         & ManiSkill         & CALVIN            \\
\midrule
FT reference (no merge)         & $66.8$            & $90.0$            & $86.5$            \\
\midrule
TIES VLM ($k{=}0.8$, $s{=}0.5$) & $1.9$ ($-64.9$)   & $64.7$ ($-25.3$)  & $0.0$ ($-86.5$)   \\
WUDI VLM (300 iter)             & $4.7$ ($-62.1$)   & $80.3$ ($-9.7$)   & $0.0$ ($-86.5$)   \\
\bottomrule
\end{tabular}
\end{table}

Table~\ref{tab:vlm_sanity} reports the penalty in success rate
relative to the same-batch FlowSDE reference. The penalties are
severe across both methods and all three benchmarks. CALVIN is
particularly diagnostic: \texttt{success\_at\_end} drops to $0\%$ for
both methods, meaning no rollout completes the entire five-step
chain. The MetaWorld penalty exceeds $-60$\,pp for both methods,
indicating that even strong consolidation strategies destroy the
language-instruction routing required to identify the active sub-task
within the 50-task suite. Only WUDI on ManiSkill stays within
$10$\,pp of the FlowSDE reference, a single positive cell out of
six, far below any usable threshold.

We conclude that static VLM consolidation, even in this favorable
setting (action expert is GT), is not viable as a standalone design.
The diagnostic also clarifies what kind of information the VLM
encodes: cross-task language-instruction grounding, which is highly
sensitive to parameter perturbation in a way that action-side
representations are not. NestedVLA therefore confines its
hypernetwork to the action expert and keeps per-task VLM storage; an
extension to also generate VLM deltas via a hypernetwork is left to
future work.

\paragraph{Full-model extension (v2).}
We use v2 to denote a possible extension in which the addressed parameter
blocks cover both the VLM and the action expert, so the hypernetwork
generates both updates for the selected task. This removes the stored VLM
values but substantially enlarges the output space and may require additional
generator capacity and metadata. We do not evaluate this extension here;
all reported results and storage measurements correspond to v1.

\section{Loss-Side Ablation}
\label{app:loss_ablation}

This appendix expands the brief discussion in \S\ref{sec:exp_ablation}
of the three knobs of the training objective in
Eq.~\ref{eq:loss}: the loss space (per-tensor standardized vs.\
absolute), an in-mask magnitude weighting $|\tau_m|^p$ that
emphasizes large-magnitude positions, and the out-of-mask penalty
weight $w_\text{out}$ that pulls $\mathcal{G}_\phi$ toward zero on
$M_m{=}0$ positions. Table~\ref{tab:loss_ablation} reports five
configurations evaluated under the $6{,}000$-step schedule of our
default architecture.

\begin{table}[t]
\centering
\small
\setlength{\tabcolsep}{6pt}
\caption{%
Loss-side ablations on top of the architecture in
Table~\ref{tab:arch_ablation}. We vary the loss space
(\textit{std}: per-tensor standardized; \textit{abs}: raw absolute),
the in-mask magnitude exponent $p$ (each loss term is multiplied by
$|\tau_m|^p$), and the out-of-mask penalty weight $w_\text{out}$. All
runs share the same $6{,}000$-step schedule. Single-seed evaluation;
the headline NestedVLA number in Table~\ref{tab:main_hypernet} is the
$3$-seed average of the same configuration.}
\label{tab:loss_ablation}
\begin{tabular}{l c c c c c c c}
\toprule
Variant & space & $p$ & $w_\text{out}$ & MW & MA & CA & Avg. \\
\midrule
default                  & std & 0 & 0.05 & 29.7 & 90.9 & 66.7 & 62.4 \\
\textit{w/o} out-penalty ($w_\text{out}{=}0$) & std & 0 & 0    & \textbf{37.9} & 88.8 & 67.7 & \textbf{64.8} \\
abs space                & abs & 0 & 0.05 & 23.0 & 87.8 & 67.7 & 59.5 \\
mag-weighting ($p{=}1$)  & std & 1 & 0.05 & 27.7 & 86.9 & 56.2 & 57.0 \\
abs $+$ mag $p{=}1$      & abs & 1 & 0.05 & 31.2 & 88.1 & 70.8 & 63.4 \\
\bottomrule
\end{tabular}
\end{table}

\paragraph{Removing the out-of-mask penalty is the most effective lever.}
Setting $w_\text{out}=0$ improves average success by $+2.4$\,pp,
the largest single-knob effect we measure. The mechanism is mechanical:
positions outside $M_m$ are masked away at inference and contribute
nothing to the deployed delta, so penalizing them at train time
forces $\mathcal{G}_\phi$ to allocate capacity to a function it does
not need. Empirically, the out-of-mask region's loss drifts to
$\sim$$20$ in standardized units when $w_\text{out}=0$, but this
drift is absorbed by the inference-time mask and does not affect
deployed performance.

\paragraph{Loss space and magnitude weighting are non-monotone.}
The other two knobs each \emph{harm} performance when applied alone:
absolute loss space costs $-2.9$\,pp and magnitude weighting at
$p{=}1$ costs $-5.4$\,pp relative to the default. Yet
\texttt{abs $+$ mag p=1} recovers most of the loss ($-0.6$\,pp). We
interpret this as evidence that absolute and magnitude weighting
function as implicit regularizers compensating for the unbounded
growth of the out-of-mask region; when the standardized loss already
provides scale normalization, neither operator is needed in
isolation. The standardized $+$ $w_\text{out}=0$ recipe cleanly
beats every other configuration we tested and is our final default.

\section{Limitations}
\label{app:limitations}

The current evaluation has four limitations that we mark as future
work. \textbf{(i) Skill-set scale.} All experiments consolidate
$M{=}3$ skills; we have not measured how training time, hypernetwork
capacity, or output quality scale as $M$ grows by an order of
magnitude. \textbf{(ii) Adding new skills.} A new skill currently
requires extending the per-task patch-embedding table, the per-task
multi-head output, and re-training the hypernetwork; we have not
demonstrated incremental or low-cost addition of an unseen skill.
\textbf{(iii) VLM consolidation.} The VLM half is kept per-task as a
top-$K$ compressed delta, contributing $\sim$$0.35\times$ of the
$1.5\times$ budget; extending the hypernetwork to also generate VLM
deltas would lower this floor but is left to future work
(App.~\ref{app:vlm_sanity} shows that na\"ive static VLM merging is
not viable). \textbf{(iv) Real-robot evaluation.} All numbers are
from simulated benchmarks (MetaWorld, ManiSkill, CALVIN); transfer to
physical hardware is not evaluated.

\section{Broader Impact}
\label{app:broader_impact}

NestedVLA reduces the per-skill storage and serving cost of multi-task
robotic policies by replacing per-task fine-tuned checkpoints with a
single shared parameter generator. Like other efficient deployment
methods, lowering this cost can amplify both beneficial uses
(low-resource personalisation, on-device multi-skill manipulation,
faster transfer between robot platforms) and risks already present
in the underlying VLA stack, such as dual-use of robotic policies in
unintended physical environments. Our hypernetwork operates entirely
on parameter deltas of an existing publicly released backbone
($\pi_{0.5}$); we introduce no new training data or labels, and the
deployed behaviour inherits the biases and failure modes of the
source FlowSDE checkpoints. All our experiments are simulated; no
physical robot is operated in this study. We see no direct path from
the parameter-generation framework to a novel misuse scenario beyond
those already discussed in the broader VLA literature.

\newpage
\section*{NeurIPS Paper Checklist}

\begin{enumerate}

\item {\bf Claims}
    \item[] Question: Do the main claims made in the abstract and introduction accurately reflect the paper's contributions and scope?
    \item[] Answer: \answerYes{}
    \item[] Justification: The abstract and \S\ref{sec:intro} state four
      headline claims---parameter-generation framing, $+41$\,pp average
      success over the strongest dynamic merging baseline, matching
      per-task fine-tuning within $1$\,pp on ManiSkill, and $1.5\times$
      storage---each backed by Tab.~\ref{tab:main_hypernet} and
      \ref{tab:per_task_ceiling} and discussed in \S\ref{sec:exp_main}.
      Limitations are made explicit in \S\ref{app:limitations}.

\item {\bf Limitations}
    \item[] Question: Does the paper discuss the limitations of the work performed by the authors?
    \item[] Answer: \answerYes{}
    \item[] Justification: \S\ref{app:limitations} is a dedicated
      section discussing four limitations: (i) the consolidation
      experiments are at $M{=}3$ skills only and we do not measure
      scaling, (ii) adding a new skill requires retraining the
      hypernetwork, (iii) the VLM half is still stored per-task,
      contributing $\sim$$0.35\times$ of the $1.5\times$ budget, and
      (iv) all evaluations are simulated.

\item {\bf Theory assumptions and proofs}
    \item[] Question: For each theoretical result, does the paper provide the full set of assumptions and a complete (and correct) proof?
    \item[] Answer: \answerNA{}
    \item[] Justification: The paper presents an empirical methods
      contribution and does not include formal theoretical results.

    \item {\bf Experimental result reproducibility}
    \item[] Question: Does the paper fully disclose all the information needed to reproduce the main experimental results of the paper to the extent that it affects the main claims and/or conclusions of the paper (regardless of whether the code and data are provided or not)?
    \item[] Answer: \answerYes{}
    \item[] Justification: \S\ref{sec:exp_setup} reports the main
      training and evaluation settings. App.~\ref{app:implementation}
      (Tab.~\ref{tab:impl_hparams}) provides the full hypernetwork
      architecture, optimizer, schedule, task weights, evaluation
      protocol, and seed counts. Baselines and their hyperparameters
      are documented in App.~\ref{app:baselines}
      (Tab.~\ref{tab:baseline_hparams}). The storage budget reported
      in Tab.~\ref{tab:main_hypernet} is derived component-wise in
      App.~\ref{app:storage}.

\item {\bf Open access to data and code}
    \item[] Question: Does the paper provide open access to the data and code, with sufficient instructions to faithfully reproduce the main experimental results, as described in supplemental material?
    \item[] Answer: \answerNo{}
    \item[] Justification: The core NestedVLA algorithm is publicly
      available at
      \url{https://github.com/WanyuGroup/NeurIPS2026-NestedVLA}, with
      installation instructions and a minimal example. The current
      release does not include the benchmark-specific training and
      evaluation pipelines, simulator configurations, or checkpoints
      needed to reproduce every reported result. The data assets we
      rely on are public: the $\pi_{0.5}$ backbone, the FlowSDE
      training pipeline, the RLinf evaluation pipeline, and the three
      benchmarks (MetaWorld, ManiSkill, CALVIN). Their citations and
      access points appear at first mention in
      \S\ref{sec:exp_setup} and App.~\ref{app:implementation}.

\item {\bf Experimental setting/details}
    \item[] Question: Does the paper specify all the training and test details (e.g., data splits, hyperparameters, how they were chosen, type of optimizer) necessary to understand the results?
    \item[] Answer: \answerYes{}
    \item[] Justification: \S\ref{sec:exp_setup} states the backbone,
      benchmarks, evaluation protocol, and main hyperparameters of the
      hypernetwork. App.~\ref{app:implementation}
      (Tab.~\ref{tab:impl_hparams}) lists every architectural,
      optimisation, and evaluation choice. Baseline configurations are
      summarised in Tab.~\ref{tab:baseline_hparams}.

\item {\bf Experiment statistical significance}
    \item[] Question: Does the paper report error bars suitably and correctly defined or other appropriate information about the statistical significance of the experiments?
    \item[] Answer: \answerYes{}
    \item[] Justification: All numbers in
      Tab.~\ref{tab:main_hypernet} and Tab.~\ref{tab:per_task_ceiling}
      are averaged over $3$ evaluation seeds with fixed initial-state
      sampling; the typical seed-to-seed standard deviation is
      $\pm 2.7\%$ (\S\ref{sec:exp_setup}). Architecture ablations
      (Tab.~\ref{tab:arch_ablation}) report $3$-seed averages with
      std.\ $\pm 0.05$ on the in-mask loss.

\item {\bf Experiments compute resources}
    \item[] Question: For each experiment, does the paper provide sufficient information on the computer resources (type of compute workers, memory, time of execution) needed to reproduce the experiments?
    \item[] Answer: \answerYes{}
    \item[] Justification: All hypernetwork runs fit on a single
      NVIDIA A6000 GPU ($48$\,GB) at $\sim$$10$\,h wall-clock per run,
      with peak memory $42.6$\,GB; full details are in
      App.~\ref{app:implementation} (Tab.~\ref{tab:impl_hparams}).
      Per-task FlowSDE checkpoints used as the merging baselines are
      reused from publicly available recipes and are not re-trained
      for this paper.

\item {\bf Code of ethics}
    \item[] Question: Does the research conducted in the paper conform, in every respect, with the NeurIPS Code of Ethics \url{https://neurips.cc/public/EthicsGuidelines}?
    \item[] Answer: \answerYes{}
    \item[] Justification: We have reviewed the NeurIPS Code of
      Ethics. The work uses publicly available simulated benchmarks
      and a publicly released VLA backbone; it involves no human
      subjects, scraped data, or sensitive information. Anonymity is
      preserved at submission time.

\item {\bf Broader impacts}
    \item[] Question: Does the paper discuss both potential positive societal impacts and negative societal impacts of the work performed?
    \item[] Answer: \answerYes{}
    \item[] Justification: A short Broader Impact statement is
      included in App.~\ref{app:broader_impact}: lower per-skill
      storage and serving cost can amplify both beneficial uses
      (low-resource personalisation, multi-skill robotic deployment)
      and downstream risks already present in the underlying VLA
      backbone (e.g., dual-use of robotic policies). NestedVLA
      inherits the biases of the source checkpoints and introduces no
      new training data.

\item {\bf Safeguards}
    \item[] Question: Does the paper describe safeguards that have been put in place for responsible release of data or models that have a high risk for misuse (e.g., pre-trained language models, image generators, or scraped datasets)?
    \item[] Answer: \answerNA{}
    \item[] Justification: The paper does not release a new high-risk
      model, dataset, or generative system. The base $\pi_{0.5}$
      backbone is publicly available with its own release terms, and
      we add only a parameter generator that targets that backbone in
      a simulated benchmark setting.

\item {\bf Licenses for existing assets}
    \item[] Question: Are the creators or original owners of assets (e.g., code, data, models), used in the paper, properly credited and are the license and terms of use explicitly mentioned and properly respected?
    \item[] Answer: \answerYes{}
    \item[] Justification: All used assets are cited at first mention:
      the $\pi_{0.5}$ backbone~\citep{pi0.5}, the
      FlowSDE training pipeline~\citep{chen2025pi_}, the
      RLinf evaluation pipeline~\citep{yu2025rlinf}, and the three
      benchmarks (MetaWorld~\citep{mcleanmeta},
      ManiSkill~\citep{maniskill}, CALVIN~\citep{calvin}). Each is used
      in accordance with its public release terms.

\item {\bf New assets}
    \item[] Question: Are new assets introduced in the paper well documented and is the documentation provided alongside the assets?
    \item[] Answer: \answerYes{}
    \item[] Justification: The released algorithm code is documented at
      \url{https://github.com/WanyuGroup/NeurIPS2026-NestedVLA}, which
      provides installation instructions, a minimal example, the scope
      of the release, and its license. The architecture, training
      schedule, and evaluation protocol are further documented in
      App.~\ref{app:implementation}.

\item {\bf Crowdsourcing and research with human subjects}
    \item[] Question: For crowdsourcing experiments and research with human subjects, does the paper include the full text of instructions given to participants and screenshots, if applicable, as well as details about compensation (if any)?
    \item[] Answer: \answerNA{}
    \item[] Justification: The paper does not involve crowdsourcing or
      research with human subjects.

\item {\bf Institutional review board (IRB) approvals or equivalent for research with human subjects}
    \item[] Question: Does the paper describe potential risks incurred by study participants, whether such risks were disclosed to the subjects, and whether Institutional Review Board (IRB) approvals (or an equivalent approval/review based on the requirements of your country or institution) were obtained?
    \item[] Answer: \answerNA{}
    \item[] Justification: The paper does not involve human subjects
      research.

\item {\bf Declaration of LLM usage}
    \item[] Question: Does the paper describe the usage of LLMs if it is an important, original, or non-standard component of the core methods in this research? Note that if the LLM is used only for writing, editing, or formatting purposes and does \emph{not} impact the core methodology, scientific rigor, or originality of the research, declaration is not required.
    \item[] Answer: \answerNA{}
    \item[] Justification: LLMs were used only for incidental writing
      and editing assistance. They are not part of the core method,
      data pipeline, or scientific contribution: the proposed
      hypernetwork is a small task-conditioned generator over
      compressed task-vector patches, with no language-model component
      beyond the frozen $\pi_{0.5}$ VLM backbone.

\end{enumerate}

\end{document}